\documentclass[preprint, 12pt]{amsart}
\usepackage[table]{xcolor}
\usepackage{times}
\usepackage{amssymb, amsmath}
\usepackage{amsthm}
\usepackage{tikz}
\usepackage{bm}
\usetikzlibrary{matrix,arrows}
\usepackage{enumitem}
\usepackage{booktabs}

\theoremstyle{plain}

\theoremstyle{remark}

\definecolor{Gray}{gray}{0.95}
\newcolumntype{g}{>{\columncolor{Gray}}c}

\begin{document}
\title{Outlier Detection in High Dimensional Data}
\author{Firuz Kamalov$^1$$^{*}$ and  Ho Hon Leung$^2$}

\address{$^{1}$ Canadian University Dubai, Dubai, UAE.}
\email{\textcolor[rgb]{0.00,0.00,0.84}{firuz.kamalov@unl.huskers.edu}}

\address{$^{2}$ UAE University, Al Ain, UAE}
\email{\textcolor[rgb]{0.00,0.00,0.84}{hl363@cornell.edu}}

\date{\today
\newline \indent $^{*}$ Corresponding author}

\begin{abstract}
High-dimensional data poses unique challenges in outlier detection process. Most of the  existing algorithms fail to properly address the issues stemming from a large number of features. In particular, outlier detection algorithms perform poorly on data set of small size with a large number of features. In this paper, we propose a novel outlier detection algorithm based on principal component analysis and kernel density estimation. The proposed method is designed to address the challenges of dealing with high-dimensional data by projecting the original data onto a smaller space and using the innate structure of the data to calculate anomaly scores for each data point.
Numerical experiments on synthetic and real-life data show that our method performs well on high-dimensional data. In particular, the proposed method outperforms the benchmark methods as measured by the $F_1$-score. Our method also produces better-than-average execution times compared to the benchmark methods.   
\end{abstract}

\maketitle

\section{Introduction}

Despite the growing amount of data that has become available for research and discovery
there remain areas where certain type of data is scarce. In the fields such as medical diagnostics, network intrusion detection, fraudulent financial transactions and many others, deviations from normal behavior, i.e. anomalies, are rare. However, often, these events are of the greatest importance. For example, it would be  extremely beneficial to determine if a person has an illness based on abnormal lab results. Determining such singular events among the overwhelmingly `normal' instances is called outlier detection. The goal of this paper is to introduce a novel outlier detection algorithm based on dimensionality reduction and density estimation that is shown to perform favorably in comparison with other state-of-the-art techniques.

In a standard outlier detection problem given an unlabeled dataset one must determine
which of the instances do not belong to the majority of the data. We implicitly assume that the majority of the data is generated through a normal process and outliers are the instances that are not generated through the same mechanism. The outlier detection problem differs from classification problems due to extreme scarcity of abnormal instances. This difference is especially acute in high-dimensional data where most of the standard classification algorithms require large datasets for optimal performance. Therefore, outlier detection problems require a different solution from classification problems. There exist various approaches to outlier detection and each with its own methodology. Most of the methods can be categorized as semi-supervised procedures that use unsupervised learning in the training stage followed by a supervised testing stage. 

In the training stage of outlier detection process unlabeled dataset is analyzed to  identify potential anomalies. A wide range of unsupervised learning algorithms have been developed for this  purpose (Kriegel et al 2010). We can divide unsupervised outlier detection approaches into three broad categories: model-based, distance-based,  and density-based algorithms. In a model-based approach the data is assumed to be generated through some statistical distribution. The parameters of the distribution (mean, variance, etc) are calculated  based on the training set. Then the instances with low probabilities are deemed to be outliers. In distance-based approaches, the idea is that the instances  that is at a great distance from the reference set are outliers. For example, a point whose $k$-nearest neighbors lie in close proximity is judged to be normal. Alternatively, a point that is  far away from its neighbors is labeled abnormal. In a density-based approach, the density of the data in the neighborhood  of a point is used to  measure the anomaly of the point. These approaches are based on the assumption that an outlier instance has a relatively low density than its neighbors. 

Outlier detection algorithms can also be categorized as global or local methods. The main difference between the two categories lies in the reference set used by an algorithm. Algorithms that use all of the data to classify a point are called global methods. For example, many of the model-based algorithms use all the data to calculate model parameters. The main drawback of global methods is that they include potential outliers in their calculations which result in biased estimates. Local algorithms use only use a portion of the  data to establish if a point is an outlier. The main motivation for local algorithms is uneven density of data. Therefore, many density-based algorithms are local in nature - using a version of $k$- nearest neighbors ($k$NN). It is an effective approach to dealing with data with highly uneven density distribution. However, the computation of $k$NNs is a time-consuming process that makes such algorithms run slower.

In the testing stage a labeled dataset is used to measure the performance of the algorithm. Since the abnormal instances make up only a small fraction of the dataset, the performance evaluation of the algorithm needs to be considered carefully (Campos et al 2015). Standard measures such as the accuracy rate are not suitable in the case where  the majority class data is overwhelmingly larger than minority class data. For example,
if 99.9\% of the data is normal, then by simply labeling all the instances as normal would yield a 99.9\%  accuracy rate even though we would miss all the outliers. Therefore, measures such as area under the ROC curve (AUC), precision and recall are more appropriate in such scenario. AUC requires the algorithm to output an outlier score
that is used to plot the true positive rate against the false positive rate. Alternatively, precision and recall can be used with binary output.

Our goal in this paper is to address the issue of high-dimensional data in the context of outlier detection. High-dimensional data can cause serious issues when we apply many of the existing outlier detection methods. In model-based methods high-dimensional data results in high degrees of freedom which makes the objects statistically indistinguishable. In distance-based methods, minor random perturbations in the features of an object can result in significant changes in Euclidean distance. Thus, a normal point can easily become an outlier. In the density-based methods high dimensionality causes data sparsity. As a result, local density of different objects start to converge which make detecting outliers more challenging. To address the issues related to high dimensional data, we propose a novel method based on principal component analysis  and local kernel density estimation. One of the main advantages of  our method is its computational efficiency. High-dimensional data requires greater computational resources. Most of the existing algorithms do not scale well to high dimensional spaces. By reducing the dimensionality of the data we are able to address the issue of data sparsity as well as  alleviate the computational load of the algorithm. As a result, we obtain a relatively fast and accurate method for outlier detection. 

Our paper is structured as follows. In Section 2, we survey the current literature in the field of outlier detection. In Section 3, we describe the details of the proposed method including principal component analysis and kernel density estimation. In Section 4, we present the results of numerical experiments performed to test the efficacy of the proposed method. We conclude with a brief summary of the paper in Section 5. 

\section{Literature}

Model-based approaches are the earliest and most commonly used methods for outlier detection. These approaches are based on the assumption that normal data is generated through a set of statistical distribution (Barnett and Lewis, 1994). Commonly used distributions  include Gaussian, Poisson, Gamma and others.  The model parameters such as mean and variance are calculated using all the data. Then the points with low probability are deemed to be outliers. Although a model-based approach can be used as a good off-the-shelf method, it has several important drawbacks. The model parameters are calculated using the data that includes potential outliers which introduces a bias in the model. Selecting the correct model can also present a challenge. In many practical applications the data may not follow a commonly used distribution. In addition, in the case
of high-dimensional data, there may not be enough data points to construct a reliable model. 

Distance-based approaches operate on the assumption that abnormal points are far from the main cluster of points. According to one early distance method (Tukey, 1977) points on the convex hull of the full data space  are deemed as outliers. Another popular distance method uses a given radius $\delta$ and percentage $p$.  Then a point $x$ is considered an outlier if at most $p$ percent of all other points have a distance to $x$ less  than $\delta$ (Knorr and Ng, 1997). Various $k$NN-based approaches have  also been proposed to determine abnormal instances. In the simplest case, the $k$NN-distance of a point is taken as its outlier score (Ramaswamy et al 2000). Similarly, the sum of distances of a point to all its  $1$NN, $2$NN, …, $k$NN can be used as an outlier score (Angiulli and Pizzuti, 2002). We also define the outlier  score of $p$ by the number of points having $p$ among their $k$NNs (Hautamaki et al,  2004).  We note that distance-based methods are flexible and do not make any specific assumptions about the structure of the data. 
However, distance methods are computationally intensive and do not work well in cases of unevenly distributed data containing multiple clusters.

Density-based approaches compare the density of data in a local region of a point relative to data density around its neighbors. Points that have low density of data relative to its neighbors are labeled as outliers. One of the popular density methods is based on local outlier factor (LOF) (Breunig et al. 2000) which tries to address the issue of comparing the neighborhood of points from areas of different densities. Many extensions of LOF have been proposed in order to improve on the original algorithm (Jin et al. 2001, Jin et al. 2006, Tang et al. 2002). In addition to the Eucleadian distance-based density approaches, other density methods have been proposed including using kernel density estimation (Gao et al, 2011; Schubert et al, 2014; Tang and He, 2017). For example, in  (Tang and He, 2017) the  authors apply KDE to the extended $k$NNs of each point. Using the resulting probabilities, the authors compare the density at a point with the densities in the extended neighborhood. 

Despite a large amount of literature devoted to outlier detection, relatively little amount of work has been dedicated  to dimensionality reduction in the current context. Aggarwal and Yu (2001) propose partitioning the space into grids and calculating the Sparsity Coefficient (SC) for each grid cell. Then the points contained in cells with negative
SC are deemed to be outliers. In angle-based outlier degree model (Kriegel et al., 2008), the authors base their approach on the assumption that if most of the points are directed in the same direction with respect to a reference point, then the reference point is an outlier. In particular, given a reference object O and some points A and B, the angle
AOB belongs to the angle spectrum of the point O. By measuring the variance of the angle spectrum we can label points as either normal or abnormal. Since angle is a more stable quantity for measuring the difference between points in high dimensions, this approach is suitable for outlier detection in high dimensional data. In another model proposed by Kriegel et al. (2009), the authors assume that $k$NNs of outliers have a lower-dimensional projection with small variance and use the distance between the point and the orthogonal subspace of the projection as the outlier score.

Dimensionality reduction is a popular tool in machine learning. It is mainly achieved through the process of feature selection. Feature selection methods can be divided into two groups: supervised and unsupervised. Supervised algorithms can be further divided into 3 groups: filter methods, wrapper methods, and embedded methods. Filter methods use an exogenous measure to rank features or feature subsets after which the most significant features (subsets) are selected (Kamalov and Thabtah, 2017). A commonly used measure for feature (subset) evaluation is information gain. Statistical measures of significance can also be used. Wrapper methods evaluate a feature subset by running a base algorithm based on the subset and using the accuracy of the resulting classifier. Embedded methods perform feature selection as part of training a classifier as in the case of Lasso regression. Unsupervised algorithms include several algorithms such as principal component analysis, discriminant analysis, and feature similarity (Solorio-Fernández et al, 2019).   For example, feature similarity methods select a feature subset consisting of features that are pairwise least `similar' where the 'similarity' is measured using various  metrics such as information gain or $\chi^2$ statistic.

\section{Methodology}
The goal of the proposed outlier detection method is to address the issue of data sparsity
in datasets with large  numb  er of features. Given a dataset with many features, the existing outlier detection algorithms require a large number of instances to perform effectively. For example, a detection algorithm that uses  Mahalanobis distance of point $x$ to $\mu$ together with the corresponding $\chi^2$ distribution would have a large degrees of freedom. As a result, it will become harder to distinguish between different points in the dataset.  In addition, most of the existing algorithms become computationally intensive when used with high dimensional data. Our method proposes a two-step procedure consisting of principal component analysis (PCA) and kernel density estimation (KDE). We use PCA to reduce the dimension of the dataset. We then use KDE to estimate the density distribution of the data which can be used to calculate outlier score of a point. 

\subsection{PCA}
Principal component analysis is an effective and widely used dimensionality reduction tool in machine learning. The goal of PCA is to transform dataset $X$ into a lower dimensional set without losing too much information. The PCA is performed by transforming the original variables into a new set of orthogonal variables, called principal components, so that the first component is in the direction of the greatest variance of $X$, and each succeeding component in turn has the maximum variance under the condition that it is orthogonal to the preceding components. Thus PCA is a change-of-basis transformation where the new basis consists of uncorrelated components. By construction, the components obtained in the beginning of PCA contain the most amount of information about the variance of the set $X$. Likewise, the  components produced in the end of PCA transformation carry the least amount of information regarding the variance of $X$. Thus, in order to achieve dimensionality reduction of the set $X$, we can drop the last few components without losing relatively too much information.

To calculate the principal components, we must first subtract the mean of each feature from the dataset to center $X$ around the origin. Then we find the covariance matrix of the data by $X^TX$. Finally, we calculate the eigenvalues and eigenvectors of the covariance matrix. The scaled eigenvectors represent the principal components and the corresponding eigenvalues represent the amount of variance of $X$ in the direction of the eigenvectors. The exact amount of information lost by dropping a principal component $c_i$ is given by $\frac{\lambda_i}{\sum_j \lambda_j}$.


\subsection{KDE}
\begin{figure}[h!]
\center
\includegraphics[scale=0.6]{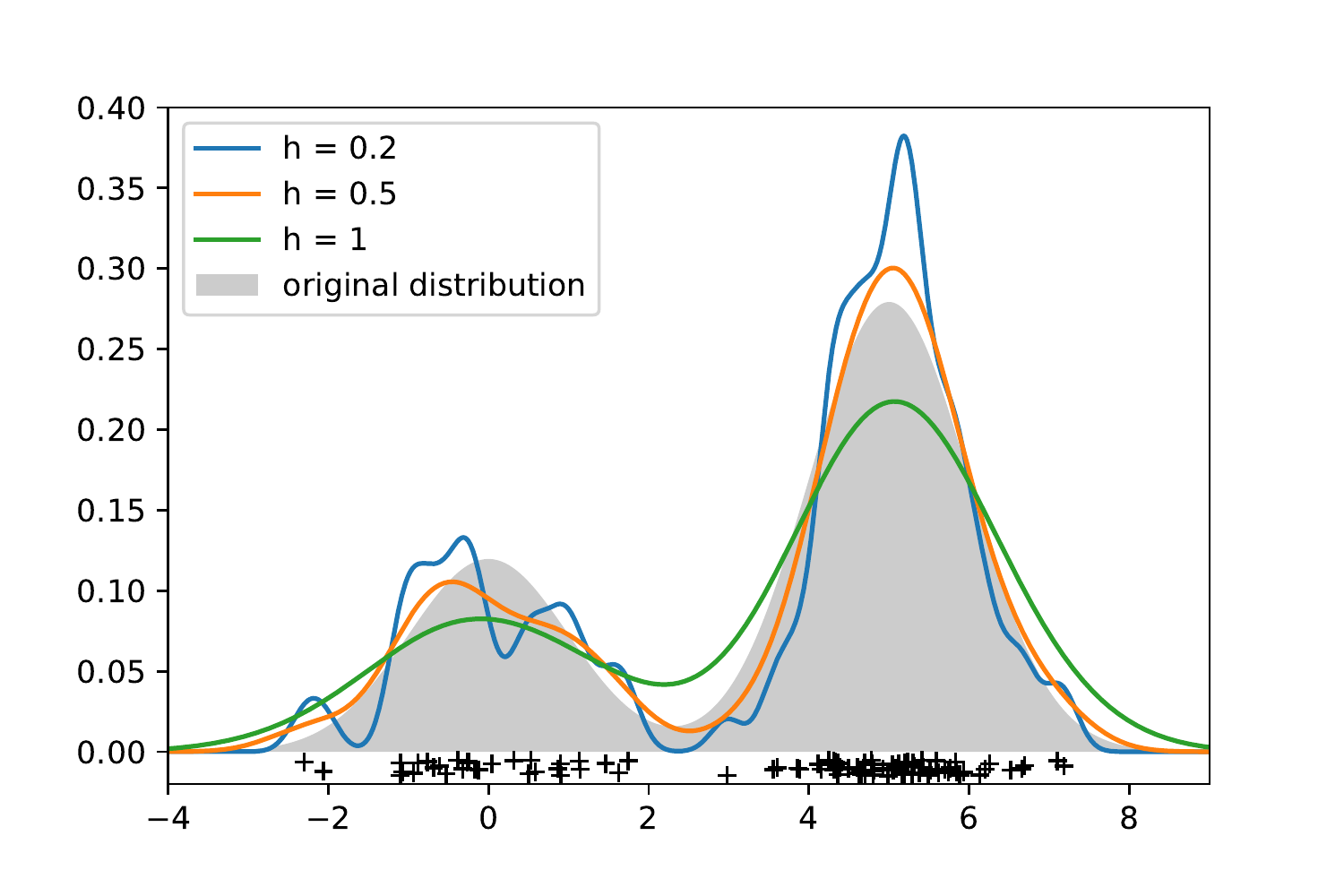}
\caption{1D kernel density estimate of a Gaussian distribution using various bandwidth values.}
\label{kde_illustr}
\end{figure}

KDE is a powerful technique for estimating the unknown density distribution based on a sample data. It has been used successfully in a wide array of applications (Gurrib and Kamalov, 2019; Silverman, 2018). Given a random sample there are two ways of estimating the true underlying distribution of the data: parametric and nonparametric. In parametric approach, we assume the general model of the density distribution and then calculate the corresponding model parameters based on the sample data. However, the model of the underlying distribution is rarely known, so parametric methods do not work well in general.
In nonparametric approach there is no assumption that the data follows some fixed distribution. Intuitively nonparametric density estimation resembles a smoothed histogram of sample data. Nonparametric methods offer more  flexibility in terms of determining the  distribution of the data. Arguably the most powerful nonparametric density estimation technique is KDE. Let $\{x_1,x_2, ..., x_n\}$ be  a sample data  consisting of $d$-dimensional vectors. According to KDE, we can estimate the true underlying density distribution $f$ of the data by

\begin{equation}
\tilde{f}(x) = \frac{1}{n}\sum_{i=1}^n K_H(x-x_i) ,
\end{equation}
where $K_H$ is the kernel function and $H$ is the bandwidth parameter matrix. Intuitively, the distribution density at a point $x$ is estimated as the average `distance' between $x$ and $\{x_i\}$ where the distance is calculated based on the kernel function $K$. The kernel function effectively transforms the Euclidean norm into a nonlinear norm.  There exist a number of kernel functions that can be used for this purpose but the most commonly used function is the Gaussian kernel. In other words, we use multivariate normal distribution as the kernel function:
\begin{equation}\label{K}
K_H(x) = \frac{1}{\sqrt{(2\pi)^d |H|}}e^{-\frac{1}{2}x^T H^{-1}x}.
\end{equation} 
Thus, we can view our kernel density estimate $\tilde{f}$ as the average over the set of Gaussian distributions centered at each sample point $\{x_i\}$,  as illustrated in Figure \ref{kde_illustr}.

The bandwidth parameter matrix $H$ controls the tradeoff between bias and variance of the resulting density estimate in addition to its orientation. For example, choosing a  diagonal matrix $H$ with small values on the diagonal we obtain a high variance and low bias estimate whose orientation is parallel to the coordinate axes. An illustration of a simple 1D KDE is shown in Figure \ref{kde_illustr}. It can be seen that a small value of  $h$ results in high variance and low bias whereas a high value of $h$ results in low variance and high bias of the density estimate.  The bandwidth matrix can be chosen in a variety of ways. In this study, we use multivariate version of Scott's rule:
\begin{equation}\label{H_val}
H  = n^{-\frac{1}{d+4}}S,
\end{equation}
where $S$ is a the data covariance matrix. Alternatively, the optimal values of the matrix $H$ can be determined via cross validation. 


\subsection{Outlier Detection Algorithm}
Our approach to outlier detection consists of two principal parts - PCA and KDE - which were described in the preceding subsections. In particular, given a dataset $X$ we first perform PCA and select the first few principal components to achieve dimensionality reduction of the data. Let us denote the reduced dataset by $X_{red}$.  Next, we use KDE to estimate the underlying distribution of the points in $X_{red}$. Note that since we have already applied PCA in the first step, the covariance matrix $S$ of $X_{red}$, which is required to determine the optimal bandwidth $H$ in KDE, will be diagonal.
Then, we select the top $K$ points with the lowest probability as the outliers. 

Our method addresses the issue of data sparsity in high-dimensional data by using PCA. In addition, our approach is computationally more efficient as it does not require any computation of $k$NNs. The  proposed method is classified as a global approach since it uses all the data. A common criticism of global approaches is that they include outlier points in their calculations resulting in biased algorithms. In order to avoid this issue we select the outliers based on the top $K$ approach instead of a threshold probability.


\section{Numerical Experiments}
In this section, we carry out a series of experiments on both synthetic and real-life data to test the efficacy of the proposed method. We benchmark the performance of our method against several other known outlier detection methods. The results of the experiments indicate that the proposed method performs as well or better than the existing popular detection methods. In addition, the proposed method is computationally more efficient compared to many of the existing algorithms as it does not require us to find the local neighborhood.   

To benchmark the performance of our method, we compare it against 5 other popular anomaly detection methods. In particular, we use Robust Covariance (Rousseeuw and Van Driessen, 1999), One-Class SVM (Schölkopf et al, 2001),  Isolation Forrest (Liu et al, 2008), Local Outlier Factor (Breunig, 2000), and ABOD (Kriegel et al., 2008) as benchmarks methods. The first 4 benchmark methods listed above were chosen due to their implementations in the popular scikit-learn library. As such these are standard outlier detection algorithms used in industry and academia. The ABOD method was initially chosen as a high dimensional benchmark method. However, due to its extremely high computational cost it is used very sparingly in this paper.

\begin{figure}[h!]
\center
\includegraphics[scale=0.6]{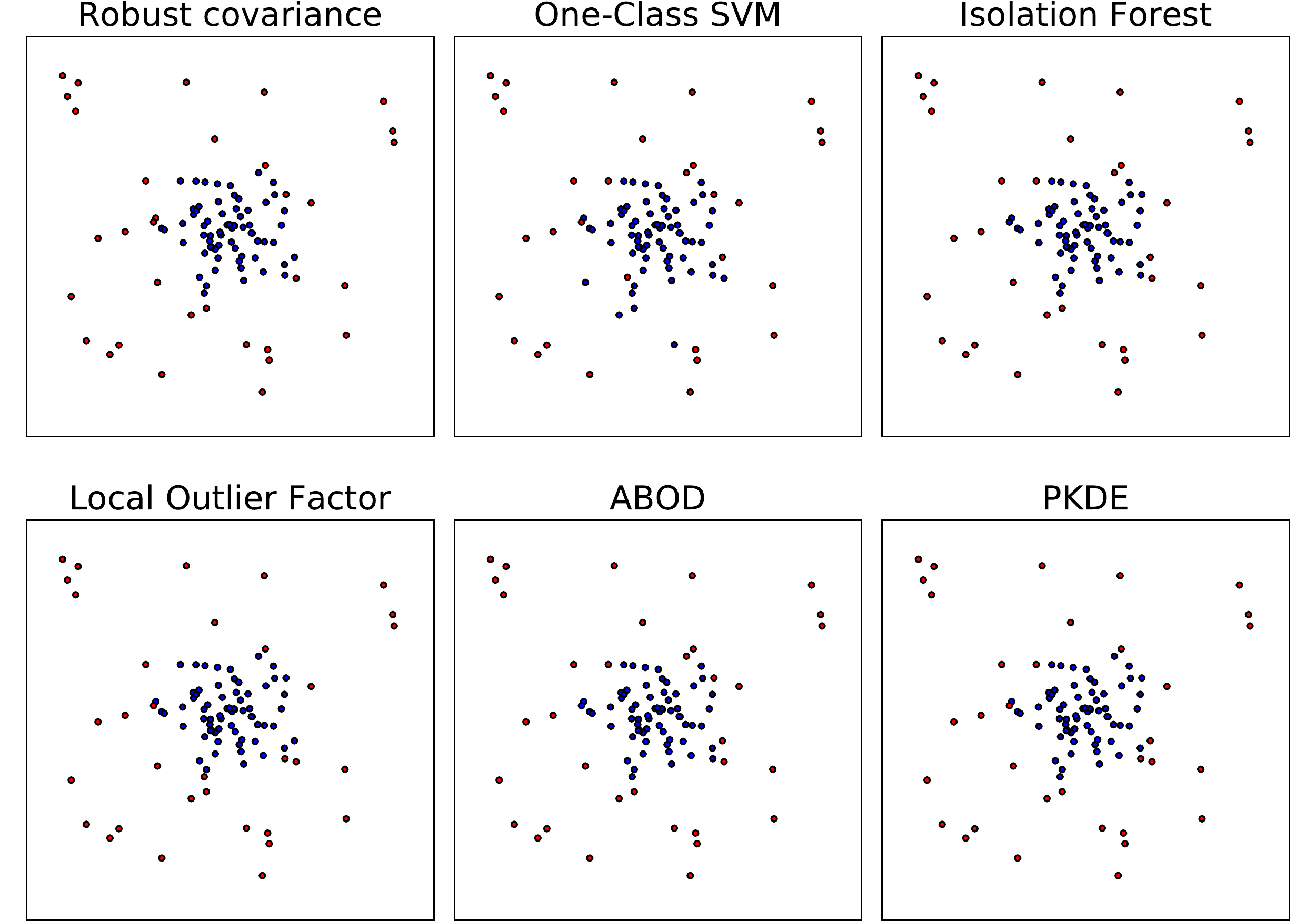}
\caption{The predicted labels by the algorithms for a 2D Gaussian dataset.}
\label{illustr_gaussian}
\end{figure}

\begin{figure}[h!]
\center
\includegraphics[scale=0.6]{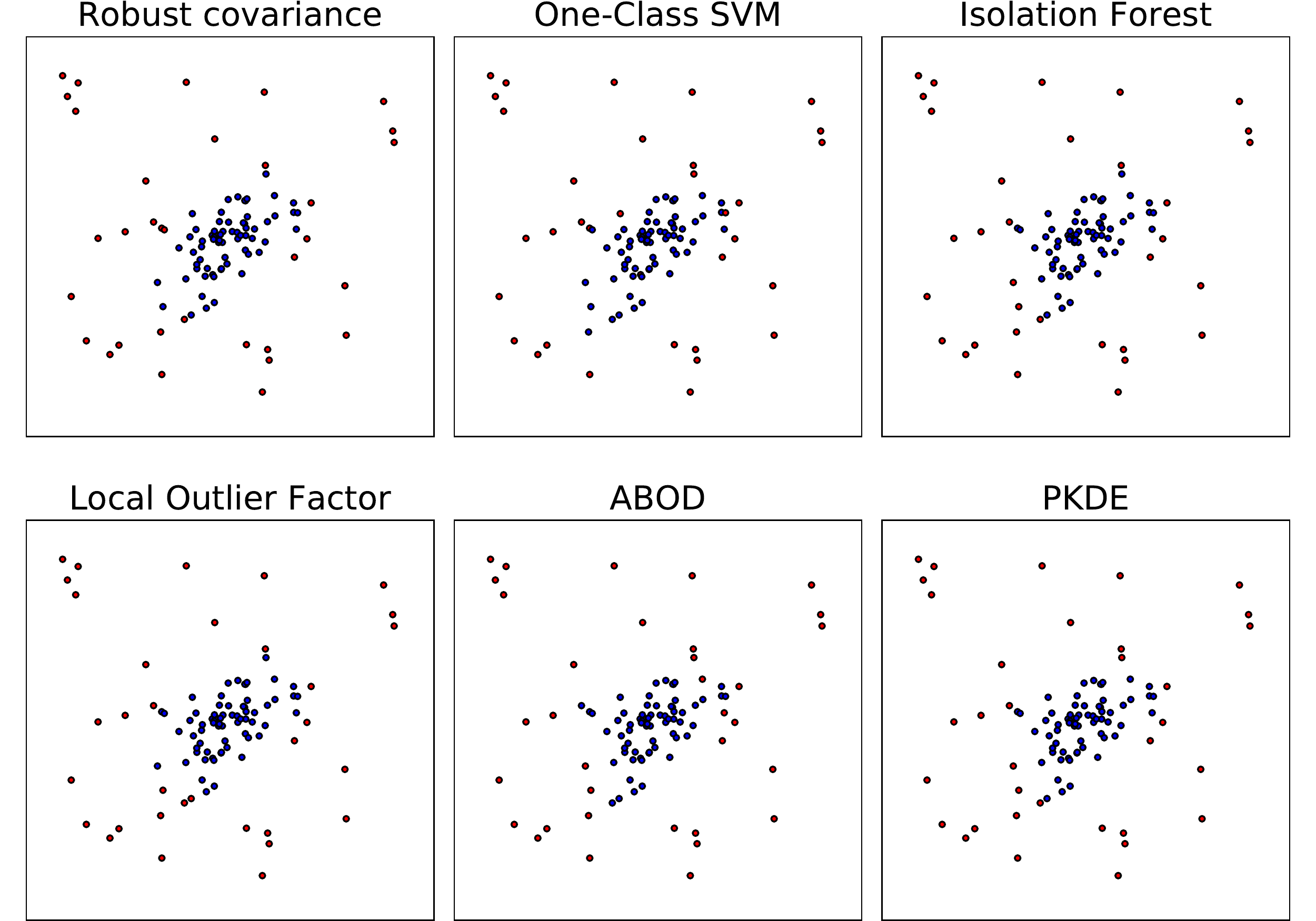}
\caption{The predicted labels by the algorithms for a 2D Gaussian dataset with nonzero covariance between the features.}
\label{illustr_gaussian_cov}
\end{figure}

\begin{figure}[h!]
\center
\includegraphics[scale=0.6]{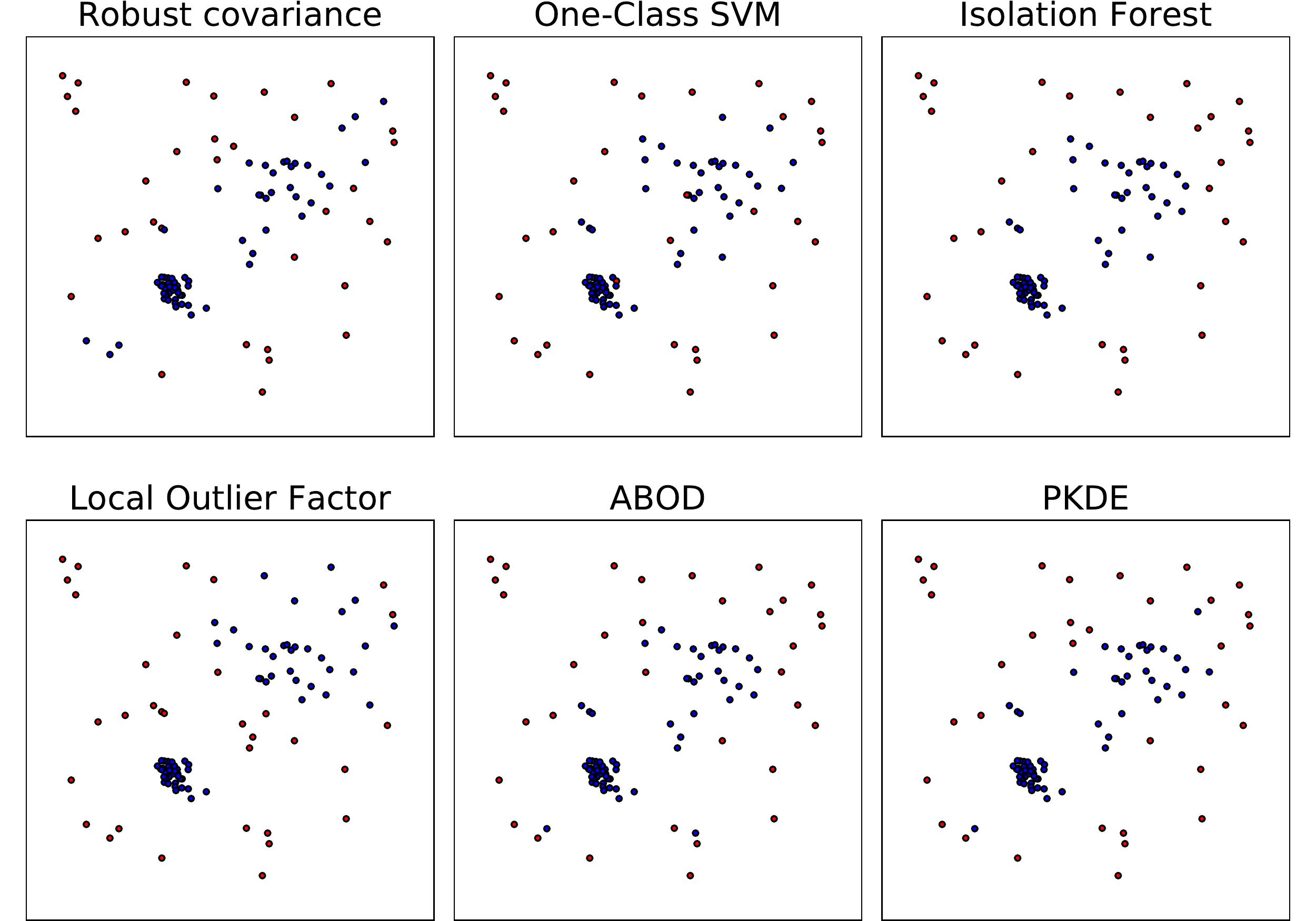}
\caption{The predicted labels by the algorithms for a dual density dataset.}
\label{illustr_dual_density}
\end{figure}

We illustrate the behavior of the algorithms in different scenarios in Figures \ref{illustr_gaussian} -  \ref{illustr_dual_density}

The key parameter used in all the methods, including our own, is the \emph{contamination} rate which is the presumed proportion of anomalies in the data. Each algorithm labels a number of points in the data as outliers based on the given contamination rate. Since, in practice, the true contamination rate is not known, we test the methods over a range of contamination values. 

In outlier detection the fraction of  abnormal instances is assumed to be vastly smaller than the normal instances. Therefore, we use precision and recall as the measure of the performance of the algorithm. More precisely, we combine precision and recall into a single value via the $F_1$-score:
$$F_1 = 2\cdot \frac{\mbox{precision}\cdot \mbox{recall}}{\mbox{precision} + \mbox{recall}}$$

\begin{figure}[h!]
\center
\includegraphics[scale=0.6]{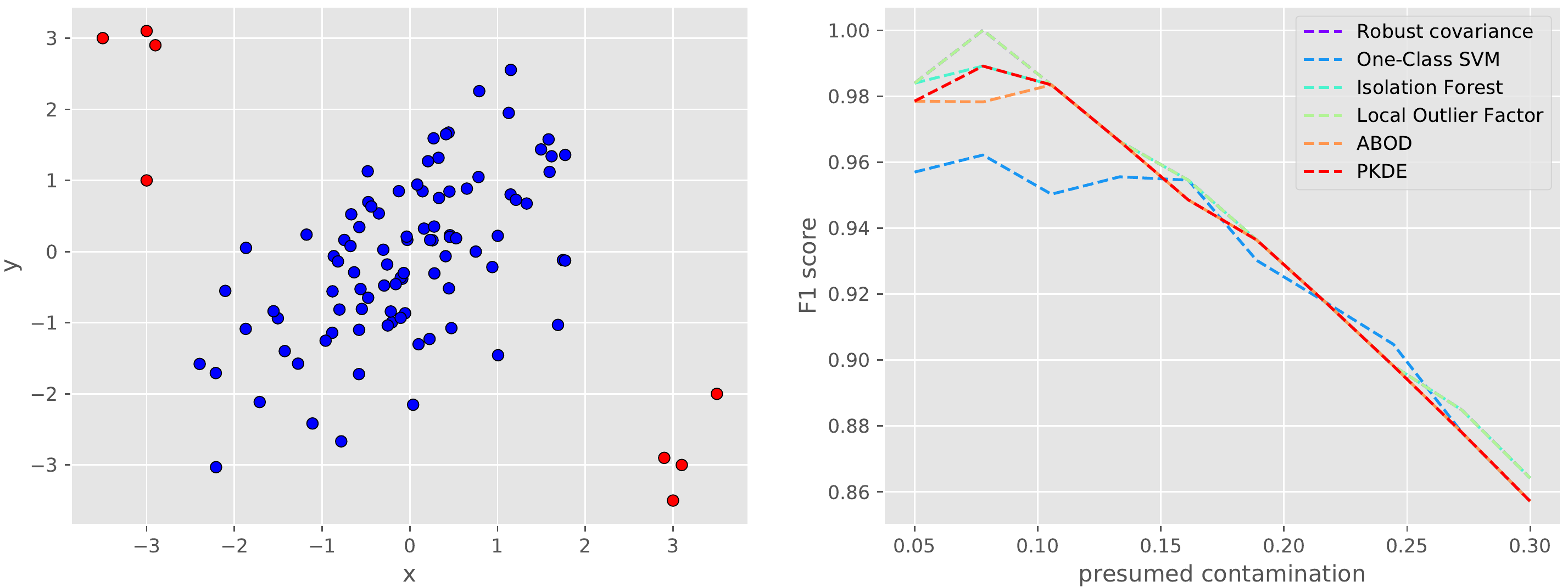}
\caption{The scatter plot of the data (left). The $F_1$-scores for different anomaly detection methods at various contamination levels (right).}
\label{data_plus_f1score}
\end{figure}

In our first experiment, we consider a simple case where normal instances are distributed according to a 2D Gaussian distribution with nonzero covariance. The abnormal instances are arbitrarily placed at a distance from the normal instances as illustrated in Figure \ref{data_plus_f1score}. We apply our method, \emph{PKDE}, along with the benchmark methods over various contamination levels and calculate the $F_1$ score as the indicator of algorithms' efficacy. The results of the experiment are presented in Figure \ref{data_plus_f1score}. We see that even in a lower dimensional setting the proposed method performs comparably to other methods. In addition, the proposed method is computationally superior to all other methods with exception of \emph{Robust covariance}.

We now apply our method to real life data. We use 4 different datasets that represent various use cases for outlier detection. The details of the datasets are given in Table \ref{info}.

\begin{table}[h!]
\centering
\caption{Experimental Datasets}
\label{info}
\begin{tabular}{llllll}
\toprule
{} &          Name &         Repository \& Target &   Ratio &      \#S &   \#F \\
\midrule
\rowcolor{Gray}
1  &         ecoli &            UCI, target: imU &   8.6:1 &     336 &    7 \\
2  &      satimage &              UCI, target: 4 &   9.3:1 &   6,435 &   36 \\
\rowcolor{Gray}
3  &  spectrometer &           UCI, target: $>$=44 &    11:1 &     531 &   93 \\
4  &     yeast\_ml8 &           LIBSVM, target: 8 &    13:1 &   2,417 &  103 \\
\bottomrule
\end{tabular}
\end{table}

\begin{figure}[h!]
\center
\includegraphics[scale=0.5]{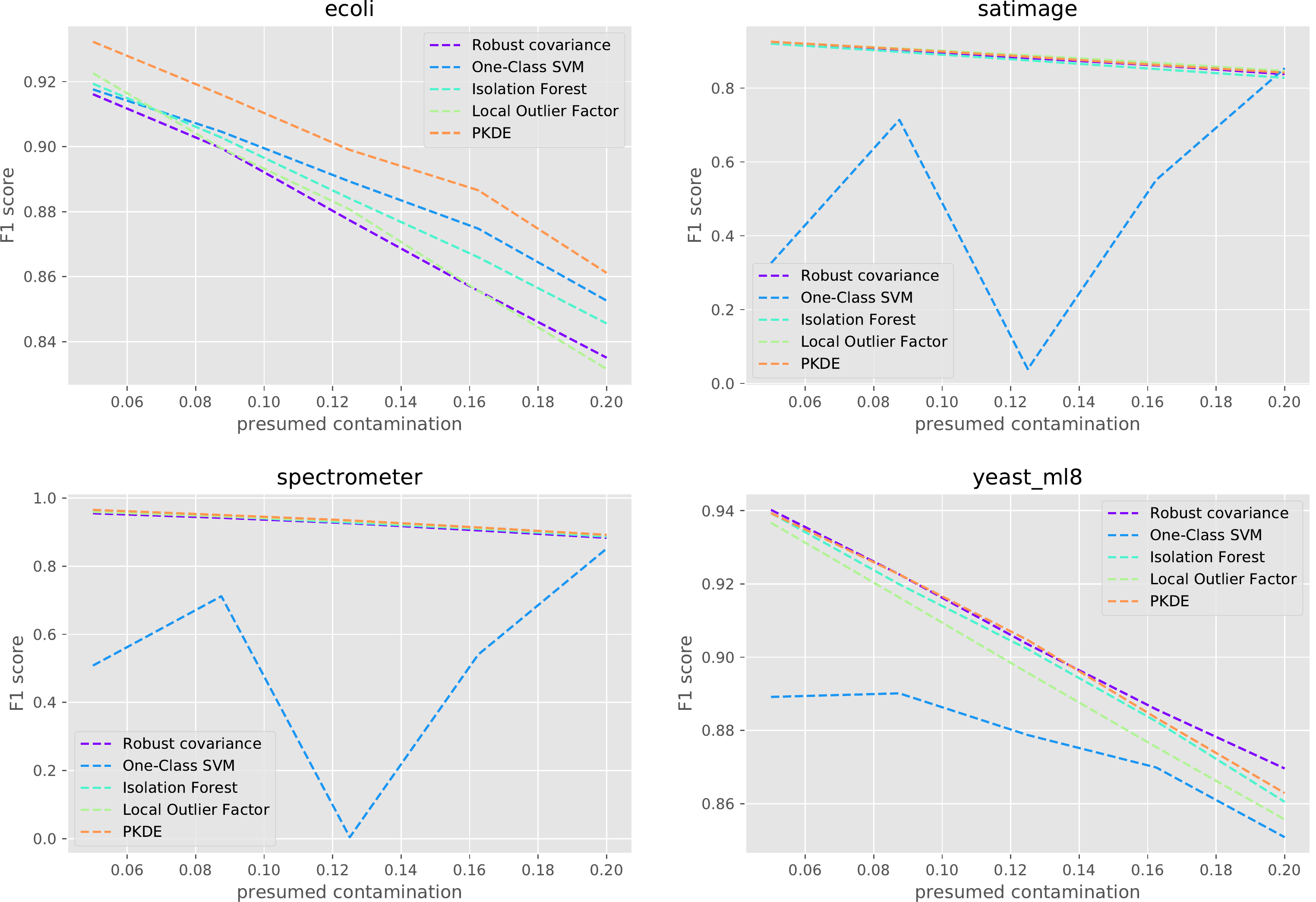}
\caption{$F_1$-score plots on different datasets with respect to the proposed and benchmark methods.}
\label{comp_exp}
\end{figure}

As can be seen from Figure \ref{comp_exp},  our method performs as well or better than benchmark methods. On the \emph{ecoli} dataset the proposed method achieves the optimal performance among all the tested algorithms. Note the PKDE produces significantly better results across different levels of presumed contamination levels. On the \emph{yeast\_ml8} the proposed method is a close second-best performer. And on the remaining 2 datasets the proposed method performs as well as any other tested algorithm. In fact, PKDE is the best performer on the \emph{spectometer} dataset by a small margin. We can thus conclude, based on the analysis of the results in Figure \ref{comp_exp}, that PKDE is overall the best performer on the 4 tested datasets.  

In addition, the execution time of the proposed algorithm is on average superior to the benchmark algorithms as illustrated in Table \ref{timings}. In particular, PKDE features the fastest execution on  \emph{spectrometer} and \emph{yeast\_ml8} datasets which also have the highest dimensions. This observation confirms our hypothesis that PKDE will performs best in the context of high dimensional data. Furthermore, PKDE achieves second and third best execution times on \emph{satimage} and \emph{ecoli} datasets respectively. Note that the \emph{ecoli} dataset has the lowest number of features. We conclude that while PKDE achieves competitive execution times on lower dimensional data, its strength lies in high dimensional data where it outperforms the other benchmark methods.

\begin{table}[h!]
\centering
\caption{Average execution times for the experiments described in Figure \ref{comp_exp}.}
\label{timings}
\begin{tabular}{lrrrrr}
\toprule
{} &  Robust covariance &  One-Class SVM &  Isolation Forest &  Local Outlier Factor &      PKDE \\
\midrule
\emph{ecoli} &           0.028124 &       0.006241 &          0.184334 &              0.000000 &  0.009372 \\
\emph{satimage} &           2.871205 &       8.369138 &          0.925531 &              1.251964 &  1.206319 \\
\emph{spectrometer} &           2.927875 &       0.112455 &          0.190598 &              0.037474 &  0.021882 \\
\emph{yeast\_ml8} &           7.187329 &       0.590327 &          0.666318 &              1.360074 &  0.541901 \\
\bottomrule
\end{tabular}
\end{table}


\section{Conclusion}

In this paper, we propose a novel method for outlier detection that addresses the issue of data sparsity in high dimensional spaces. First, we apply PCA to reduce the dimension of the data. Then we use KDE to model the density distribution of the data and select the top $K$ points with the lowest probability as outliers. The results of numerical experiments show that the proposed method performs well relative to other benchmark methods. In addition, the proposed methods is computationally lighter than many of the existing methods. As a result our method offers a relatively fast and accurate tool for outlier detection.  


\end{document}